# Social Behavioral Phenotyping of Drosophila with a 2D-3D Hybrid CNN Framework

Ziping Jiang[1], Paul L. Chazot[2], M. Emre Celebi[3], Danny Crookes[4] and Richard Jiang[1]
[1]School of Computing and Communication, Lancaster University, Lancaster, UK
[2]Deparment of Biosciences, Durham University, Durham, UK
[3]School of Computer Science, University of Central Arkansas, Conway, USA
[4]Computer Science, Queens University Belfast, Belfast, UK

Corresponding author: Richard Jiang (e-mail: r,jiang2@lancaster.ac.uk)

This work was supported in part by the EPSRC grant (EP/P009727/1).

**ABSTRACT** Behavioural phenotyping of drosphila is an important means in biological and medical research to identify genetic, pathologic or psychologic impact on animal behviour. Automated behavioural phenotyping from videos has been a desired capability that can waive long-time boring manual work in behavioral analysis. In this paper, we introduced deep learning into this challenging topic, and proposed a new 2D+3D hybrid CNN framework for drosphila's social behavioural phenotyping. In the proposed multi-task learning framework, action detection and localization of drosphila jointly is carried out with action classification, and a given video is divided into clips with fixed length. Each clip is fed into the system and a 2-D CNN is applied to extract features at frame level. Features extracted from adjacent frames are then connected and fed into a 3-D CNN with a spatial region proposal layer for classification. In such a 2D+3D hybrid framework, drosphila detection at the frame level enables the action analysis at different durations instead of a fixed period. We tested our framework with different base layers and classification architectures and validated the proposed 3D CNN based social behavioral phenotyping framework under various models, detectors and classifiers.

## I. INTRODUCTION

SOCIAL behavior analysis has a significant role in comprehension of gene expression of laboratory animals. One of the most common laboratory animals, Drosophila Melanogaster, also known as fruit flies, can exhibit a wide range of complex social behaviors though it has only 105 neurons. It also has a high frequency of social interaction, which makes it an ideal model for phenotype analysis.

However, manual phenotype analysis by naked eyes is an arduous task that requires professional knowledge and great labor. Meanwhile, the dependence on human perception sometimes introduces errors and lacks objectivity. Automated phenotyping [1] using machine learning techniques is then a sought-after capability to enable the processing of large amount videos in biologic and medical research.

Mice and drosophila behavioural phenotyping has been successfully reported [1-4]. These tracking systems typically start with feature detection for each object. Those features are then used for detecting social behaviors with statistical methods, such as support vector machine (SVM) and hidden Markov model (HVM). However, such pipelines are not transferable since they are highly dependent on the tracking system, which is often designed for a particular task with specific inputs and outputs.

The behavioral analysis of fruit flies [3-4] is somehow challenging due to two aspects. First, the legs of a fruit fly is tiny and hard to track; Second, the activities of fruit flies are very fast, often happening in several frames. These critical challenges not only make it hard for naked eyes to identify actions, but also drive the computer-based analysis to a new demand of more sophisticated methods.

In recent years, deep learning has achieved state-of-the-art results in various fields. The success of convolutional neural network (CNN) based methods for image analysis has paved the way for human action detection and recognition. As long as there is enough annotated data, a well-designed deep learning architecture can be used for various tasks with satisfactory results. Inspired by the breakthroughs in the domain, in this work, we bring deep learning in the pradigm of drosophila's behavioural analysis, and propose an end-to-end deep learning-based method for detecting and localizing social behaviors of fruit flies.

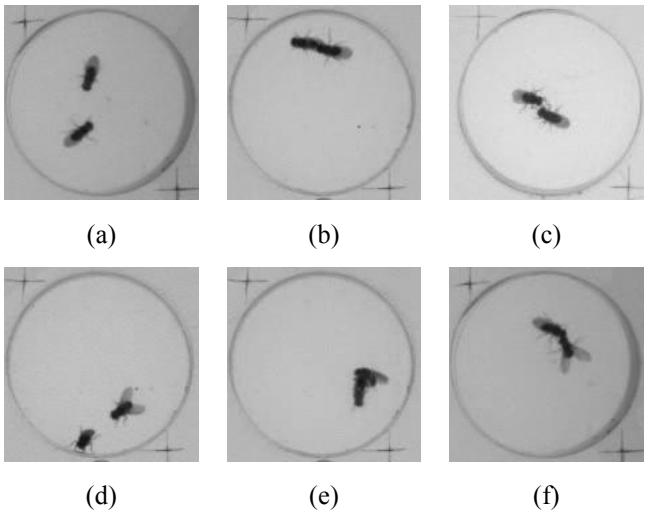

FIGURE 1: Examples of fly behaviors. (a) Normal flies without social behavior. (b) Hold behavior between flies. (c) Tussle behavior between flies. (d) Wing threat behavior of a fly. (e) Confusable tussle behavior (against hold) (f) Confusable tussle behavior (against wing threat.)

To the best of our knowledge, we are the first to explore deep learning methods in social behavior detection of multiple laboratory animals. There are several differences between action detection of animals and humans:

1) The duration of an animal behavior, sometimes lasting for only a few frames, is much shorter than that of humans, which means more attention should be paid to short-context;
2) Most videos of laboratory animals are in lower resolution, thus the feature extraction module should be carefully designed to avoid frame-wise information loss;
3) Different behaviors are more easily confused than those of humans (as shown in Fig1). A bounding box that has high intersection of union (IoU) with the ground truth box in object detection or action detection could be recognized correctly, while in behavior detection tasks they are more likely to be misclassified. For example, if only part of the behavior area is contained in the bounding box, it might be detected as a single behavior instead of social behavior.

To address the above challenges, in our work, we propose a new 2D+3D hybrid CNN framework for phenotyping.

In our method, we first use a 2D CNN for feature extraction and then use a 3D CNN for spatiotemporal information so that we can maintain a balance between spatial and temporal information. At each frame, the current frame plus adjacent frames ($2k + 1$ in total, referred as the bout length) are fed into base layers for feature extraction. After that, a Spatiotemporal Region Proposal Network (SRPN) is used to generate proposal tubes, followed by convolutional 3D layers and fully-connected layers for action classification. To boost the training process, we add a classifier for super-category detection, and the output is fed back to the action classifier to improve accuracy. For the marginal frames of a video, we extend the edge frame of the marginal side to obtain a fixed length of input so that the detection process can be performed in every frame of the input video. We train and test our method on the Fly-vs-Fly dataset [3].

We adopt VGG model and ResNet as base layers. 4 different bout lengths are applied so that the effect of utilizing spatiotemporal features for frame-wise classification can be clearly observed. We also tested architectures with and without the parent category classifier to compare the impact of the additional classifier.

To our best knowledge, our work is the first one hat combines 2D and 3D CNN for frame-wise action detection of laboratory animals. Briefly, the contributions of our work are:

1) We leveraged deep learning for Drosphila's behavioural phenotyping, and proposed a 2D+3D hybrid framework for frame-wise action detection of easily-confusing behaviors of laboratory animals. The features are extracted via 2D CNN while actions are located and classified by 3D CNN.
2) A new measurement is proposed to test the validity of bounding tube to reduce the probability of misclassifying.
3) Based on the proposed tube, a computational effective pooling layer that pools tubes with different sizes into a tube with a fixed size for further feature fusion as well as detection.
4) A super category classifier is proposed to boost the training process. An overall mAP of 63.7% is achieved.

In the following sections, section II summarizes related work, section III presents the proposed methodology, section IV gives the experimental validation and section V concludes the whole work.

## II. RELATED WORK

### A. STATISTICAL METHODS FOR ACTION DETECTION

Common approaches for animal social behavior analysis use frame-by-frame classification based on features extracted by computer vision methods. Once features are extracted, statistical learning methods are often used for detecting actions. Kabra et al. [4] developed an intuitive interactive system, Janelia Automatic Animal Behavior Annotator, to annotate laboratory animals like mice, fruit flies and larvae. Dankert *et al*. [1] computed 25 features, such as location of body, orientation of flies, along with nearest neighbor comparison for drosophila action detection. Eyjolfsdottir *et al* [3] introduced a fly-vs-fly dataset as well as a fruit fly tracking system which can extract vital features for action detection. They compared sliding window SVM against structured output SVM detectors and found that the former method outperformed their counterparts. Jhuang et al [5] presented an automated behavioral phenotyping system and used a SVMHMM [6] for detecting action of single housed mice. Xavier [2] introduced a dataset (CRIM13) containing social behavior between mice as well as a tracking system for his dataset. They use boosting and auto-context on sliding windows for action detection on their dataset.

In recent years, statistical learning methods are also applied to human activity recognition [7]. Yamato et al. adopt Hidden Markov Models for human action representation [8]. Hoai [9] proposed multi-class SVM method for video segmentation and action recognition in video. Shi [10] presents a discriminative semi-Markov model are used for segmenting human actions.

### B. OBJECT DETECTION

Neural networks with deep structures, which are known as deep models, have a long history and were popular in 1980s and 1990s [11]. However, due to the limitation of datasets and computational power, they fell out of fashion in the 2000s. Recent years, with the emergence of large annotated datasets, such ILSVR [10], PASCAL VOC [12] and the development of high-performance computing techniques such as GPUs and processor clusters, deep models have proven to be effective in many proposed models, such as VGG16 [13], ResNet [14], etc.

For large scale image object detection task, Girshick et al [15] introduced R-CNN, an inspiring two-stage architecture by combining a proposal detector and region-wise classifier. SPP-Net [16] and Fast R-CNN [17] are then introduced with the idea of region-wise feature extraction which significantly speeds up the overall detector. Girshick et al. [18] proposed a Regional Proposal Network, which is almost cost free by sharing convolutional features with the detection network, for object bounds prediction. A multistage detector Cascade R-CNN [19] is then proposed which improves the accuracy of detection by setting increasing IoU thresholds for a sequence of detectors.

One stage object detection, as an alternative architecture, is also popular due to its computational efficiency. YOLO [20] is implemented with an efficient backbone network and enables real time object detection. SSD [21] uses multiple feature maps at multiple resolution to cover objects with different scales and detects objects in a similar way to RPN [17]. The downside of faster one-stage detectors is that their accuracies are below most two-stage architectures. However, RetinaNet [22] achieved better results than most two-stage object detectors by addressing foreground-background imbalance in dense object detection.

TABLE 1: Notations

| Symbol | Description |
|---|---|
| $L$ | Length of input video clip. (Fig2) |
| $W$ | Width of input video clip. (Fig ) |
| $H$ | Hight of input video clip. (Fig ) |
| $L - 2k$ | Number of predictions made for frames per batch. |
| $k$ | Length of expansion per side. |
| $n$ | Number of bounding boxes per anchor. |
| $K = 2k + 1$ | Bout Length. |
| $Ncls$ | proposed anchors per bout. |
| $Nbatch$ | proposed anchors per batch. |
| $C$, | Number of channels in feature map. |
| $GT_{i,k}$ | Ground truth box of object $i$ in frame $k$. |
| $Anchor_j$ | $j$th proposed anchor box in a bout. |
| $ANK_{i,j,k}$ | ......rate of anchor box $j$, Ground truth box $i$ in frame |
| $AA_{i,j}$ | ......average of $ANK$ measure in all frames, used for anchors. |

### C. ACTION DETECTION AND CLASSIFICATION

Academic research in action recognition has made great progress in recent years [23]. Karpathy et al. [24] studied the performance of CNN and found that a CNN architecture is capable of action recognition in large scale video. Ng et al. [25] adopt a CNN for feature extraction, followed by a LSTM for action classification. Simonyan et al.[26] proposed a two stream ConvNet, consisting of separate networks for frame and optical flow, that incorporates spatial and temporal networks. Ji et al. [27] proposed CNN based human detector and head tracker. Tian et al. [30] proposed a simple but effective 3D CNN architecture for video classification.

Action detection is a more challenging problem than action recognition [28]. Before the deep learning era, most proposed methods were top-down based approaches. Ke et al. [29] match event models to over-segmented spatiotemporal volumes for event detection in crowded videos. Tian et al. [30] generalize deformable part models from 2D images to 3D spatiotemporal volumes to study their effectiveness for action detection. Oneata et al. [31] and Desai et al. [32] proposed sliding window-based approaches.

The success of CNN based methods for image classification paved the way for their use for action detection in videos. Gkioxari et al. [33] used CNN for feature extraction in candidate regions and used SVM to detect when and where an action is performed. Weinzaepfel et al. [34] and Peng et al. [35] proposed methods that first detect action proposals and associate actions across frames to determine true action. Saha et al. [36] adopt SSD to perform online spatiotemporal action localization in real-time. Jin et al. [37] present a sub-action descriptor for detailed action detection with multi CNN. However, those methods treat spatial and temporal features of a video separately; thus, the temporal consistency in video is not well explored in the network. Hou et al proposed T-CNN [38] and ST-CNN [39], which exploit 3D CNN for video action detection in an end-to-end model. In their work, they generalized R-CNN from 2D to 3D by using a Tube-of-Interest pooling layer. The proposals are then linked into larger tubes for action detection.

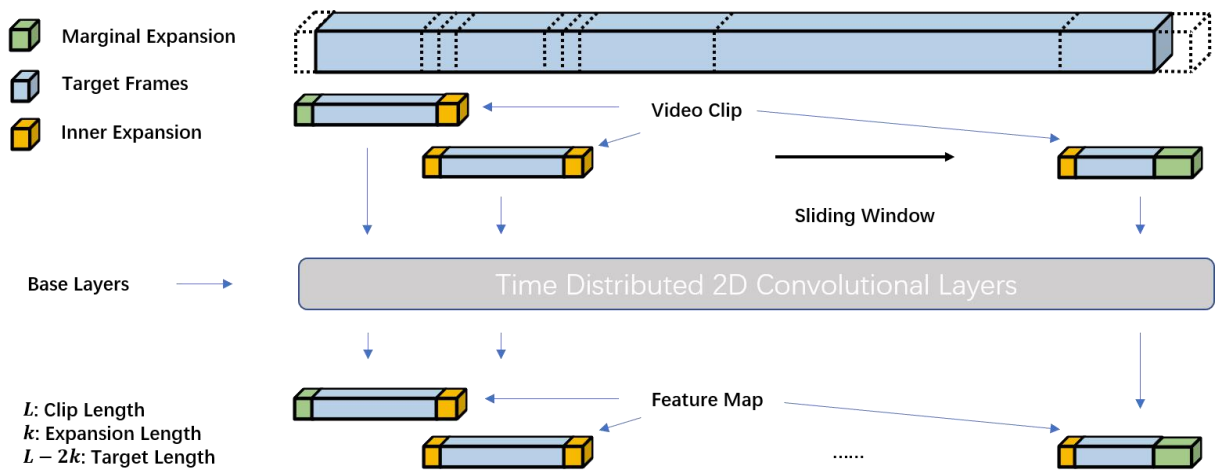

FIGURE 2: Illustration of sliding window. The blue tube at the top is an original video. Video is cut into clips with fixed length (for the first and last clip, the edge frame is extended to maintain a fixed length). A video clip is then fed into base layers for feature extraction. Notice that since the base layers are time distributed, the length of its input and output are the same but with different width and height. The feature maps are then fed into SRPN and behavior detector for classification.

Even though our task is different from previous, since our prediction is made at frame level while most action detection models are used for recognizing actions in a longer video, those tasks are still inspiring.

## III. METHODOLOGY

In this section, we introduce the structure and details of our model for action detection and location. Since most social behaviors of fruit flies may only last for a few frames, common approaches for action detection, which make prediction based on a fixed length video clip, are not suitable for our task. Instead, our system makes predictions for each frame using the information in adjacent frames. The entire system consists of two parts: base layers and classification layers. Base layers are responsible for feature extraction in each frame. The classification layers concatenate features in different frames, adopting a spatiotemporal region proposal network (SRPN) to locate and classify objects in the current frame.

### A. BASE LAYERS

For the feature extraction task, both 2D-CNN and 3D-CNN have been proven to display good performance. Although 3D CNN architecture has better performance in spatiotemporal feature learning [6] it will also result in confusion of information between time and space dimensions. It is not problematic in human action detection task. However, compared with humans, fruit flies are smaller and have faster movement and higher action similarity, which means without a carefully designed feature extraction process, the classifier can easily confuse behaviors of fruit flies. Thus, we use 2D CNN for feature extraction at the frame level.

As mentioned above, the classification task is performed for each frame and the classifier input is the features of the current and adjacent frames, which means features of a single frame may be used for K = (2k + 1) times, where K is the bout length and k is the number of adjacent frames of current frame picked for classification. To reduce the computational cost of training, we feed a clip of N frames into the base layers for feature extraction each time and then apply a sliding window for training the classifier. Once the feature map of the input clip is extracted, the sliding windows pick features of K frames centered on the current frame and make predictions.

How to deal with marginal frames is always an issue for most action detection tasks. In our work, we extend k frames of the previous batch to the beginning of the current clip and k frames of the following clip to the endpoint of the current clip to keep continuity. If the input clip is the first or last clip of a video, we extend the edge frame to maintain a fixed length of input. This process occurs during the data preprocessing stage; thus, it does not affect the structure of our model.

TABLE 2: Base Layer Structure

| block | size | VGG16 | ResNet |
|---|---|---|---|
| input | 288 | | |
| Conv1 | 288 | [3 × 3, 64] ×2 | [7 × 7, 64], stride 2 |
| | | max pooling | |
| Conv2 | 144 | [3 × 3, 128] ×2 | 3 × 3 max pool, stride 2 |
| | | max pooling | [3 × 3, 64] ×4 |
| Conv3 | 72 | [3 × 3, 256] ×3 | [3 × 3, 128] ×4 |
| | | max pooling | |
| Conv4 | 36 | [3 × 3, 512] ×3 | [3 × 3, 256] ×4 |
| | | max pooling | |
| Conv5 | 18 | [3 × 3, 512] ×3 | [3 × 3, 256] ×4 |
| | | remove max pooling | |
| Conv6 | 18 | None | [3 × 3, 256] ×4 |

Denote the size of a batch of input frames by $L \times H \times W \times C$, where $L$ is the number of frames in a clip; $H$, $L$, $C$ are height, width and number of channels respectively. The first and last $k$ frames of a clip are used for feature extraction. Thus, the classification of behaviors only is performed on the middle $L - 2k$ frames. $H$ and $L$ depend on the structure of base layers.

In our work, we apply modified VGG, ResNet as base layers. For the VGG model, we remove the last max-pooling layer and the last block; for the ResNet, we adopt an 18 layer architecture and remove the last block, as well as the max-pooling layer at block 4 and then retain convolutional layers in block 4. Fully connected layers of both models are removed.

Since the architecture of base layers is changed, we re-train the different layers. Thus, the down sampling scale of base layers in our system is fixed to 8. The size of the feature map is $L \times H/8 \times W/8 \times C^t$, where $C^t$ is the number of channels of the feature map and depends on the base layers. Details of the base layers are listed in Table 2.

## B. CLASSIFICATION LAYERS

The classification layers comprise three modules. All the modules share the 3D convolutional layers which are used to extract spatiotemporal information. The first module is a 3D convolutional network that proposes spatiotemporal regions, the latter two modules are two classifiers, which share several convolutional and fully connected layers that use the features extracted by the first module. They output parent category, a specific category of a 3D spatiotemporal region proposal, respectively. In our system, the parent category classifier is used for boosting the training process and improving the accuracy of the action detector. The structure will be discussed below.

### 1) Spatiotemporal Region Proposal Network

Inspired by R-CNN, we proposed a spatiotemporal region proposal network, which is also an anchor-based method. The difference is that we generalized R-CNN from 2D to 3D and the proposals are proposed for both spatial and temporal dimension. The SRPN takes the feature map as input, and outputs a set of video tube proposals, each with an "actionness" score. Three 3D convolutional layers with same padding are first applied to model the spatiotemporal information of the input video clip, followed by two parallel convolutional layer with valid padding. The two parallel layers are a spatiotemporal box-regression layer and a spatiotemporal box-classification layer, respectively.

In Faster R-CNN, a bounding box is centered at the sliding window, with different scales and aspect ratios. We directly adopt the settings for our bounding box at the spatial dimension. For the temporal dimension, we set a fixed bout length, as mentioned above, $2k + 1$. Thus, each bounding box can select a spatiotemporal region for detection. Notice that, unlike the image, which can be cropped and reshaped without loss of key information, cropping the video clip may lose temporal information if we apply max-pooling at its temporal dimension. It may result in a reduction of accuracy in our task, since different behaviors have high similarity in our task. To avoid the problem, we set a fixed bout length and train the classifier for different bout lengths separately.

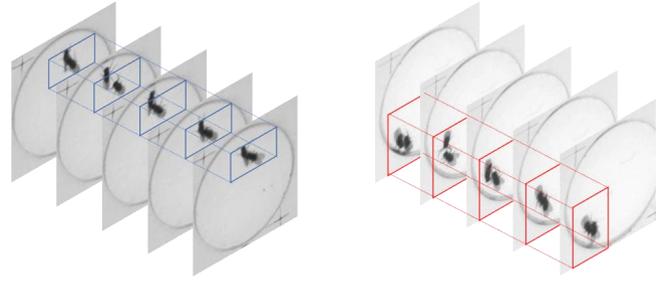

FIGURE 3: Illustration of how *AA* measure used for proposal selecting. (a) and (b) are both a bout with length of 5. (a) The ground truth boxes in frame 2 to 4 is not fully contained in blue bounding boxes. Thus, the $ANK_{i,j,2}$, $ANK_{i,j,3}$ and $ANK_{i,j,4}$ are set to be 0. Since the *AA* measurement is the mean of all five *ANK* in the bout, the blue bounding box may have a *AA* less than 0.7, thus, it is set to be negative. (b) Ground truth regions all frames are subset of the red bounding box, thus, $ANK_{i,j,k}$ are calculated according to Eq.(1). If the red bounding box has a *AA* higher than 0.7, it will be set to positive.

The spatiotemporal box classification layer(cls) has a kernel size $(2k + 1, 1, 1)$ and n channels. For each *bout* of length $(2k + 1)$, the output size from the convolutional layer is $1 \times H/4 \times W/4 \times n$. Here, $n$ denotes the number of pre-set anchors. Thus, the output of SRPN is $(L - 2k) \times H/4 \times W/4 \times n$. Each output point of SRPN is an "actionness" score, which measures the probability that the bounding box corresponds to a valid action. Similarly, the spatiotemporal box regression layer (reg) has an output of $(L - 2k) \times H/4 \times W/4 \times 4n$. That is, at each point of the feature map, there are 4n outputs encoding the coordinates for n boxes.

Most proposed method for object detection set anchor labels according to IoU overlap with ground box. However, since our method is designed for confusing behaviors detection, anchors with low IoU overlap may result in a decrease of accuracy of our model. Key information may lost if only part of ground truth box is in a proposed anchor, which may led to a confusion between behaviors (Fig). To solve the problem, define a different measure of ground truth box in a bout of frames:

$$ANK_{i,j,k} = \begin{cases} 0, & GT_{i,k} \nsubseteq Anchor_j \\ \dfrac{S(GT_{i,k})}{S(Anchor_j)}, & GT_{i,k} \subseteq Anchor_j \end{cases} \quad (1)$$

where $GT_{i,k}$ stands for the ground truth region of object *i* in *j*th frame of the bout, $Anchor_j$ denote for the region of *j*th proposed anchor box.

$$AA_{i,j} = \sum_{k=0}^{K} ANK_{i,j,k}/K \qquad (2)$$

We assign a binary class label to each anchor at each frame-centered bout. For each ground truth box, two kinds of anchors are assigned to be positive, (i) the anchor with highest *AA* overlap, or (ii) an anchor that has an *ANK* overlap higher that 0.7 with any ground truth box. Notice that higher restrictions can result in a reduction in the number of positive anchors, but such setting can improve the accuracy of action detector, which will be discussed in the next section. Then we assign a negative label to an anchor if its *ANK* overlap is lower than 0.3 for all ground truth box.

*2) Spatial Pooling Layer*

With settings listed above, we can then define the loss function of the SRPN. The loss function of a batch of frames is defined as:

$$L(\{\{p_i\}, \{t_i\}\}) = \frac{1}{N_b}\sum_{m}[\frac{1}{N_{cls}}\sum_{i} L_{cls}(p_i^m, p_i^{m*}) + \lambda \frac{1}{N_{reg}}\sum_{i} p_i^{m*} L_{reg}(t_i^m, t_i^{m*})] \qquad (3)$$

where *m* is the index of a bout, $N_b = L - 2k$ is number of bouts in a batch. $p_i^m$, $p_i^m{}^*$ denotes for the predicted probability of spatiotemporal anchor *i* of bout *j* contains an behavior and the ground truth label of the anchor, respectively. Similarly, $t^{m_i}$ and $t^{m_i*}$ are predicted coordinates and ground truth box of anchor *i* of bout *j*. The term $p^{m_i*}L_{reg}$ means the regression loss is activated only for positive anchors. Due to the restrictive conditions of setting an anchor to be positive, there are only a fewer positive anchors in a Bout (approximately 12 to 20 per Bout with default setting of our implementation). To avoid sample bias, we set the mini-batch size to be $N_{cls}$ to be 32.

Since the prediction of localization is frame-wise, we adopt the settings for bounding box regression as FRCNN:

$$\begin{aligned} t_x &= (x - x_a)/w_a, & t_y &= (y - y_a)/h_a, \\ t_w &= log(w/w_a), & t_h &= log(h/h_a), \\ t^*_x &= (x^* - x_a)/w_a, & t^*_y &= (y^* - y_a)/h_a, \\ t^*_w &= log(w^*/w_a), & t^*_h &= log(h^*/h_a), \end{aligned} \qquad (4)$$

where *x,y,w* and *h* are for the box's upper left corner coordinates and its width and height. Variables *t*, $t_a$ and $t^*$ stands for the predicted box, anchor box and ground truth box respectively. Notice that the prediction of our bounding box regression layer and bounding box classification layers are for single frame but it takes features of serval frames into consideration.

We train our SPRN end-to-end by back propagation and stochastic gradient descent (SGD)[29]. Since the input of our system is a video clip with fixed length of images, each mini-batch contains $N_{batch} = (L - 2k) \times N_{cls}$ anchors, where $L - 2k$ is the number of bouts in a batch and $N_{cls}$ is the number of proposed anchors in a bout. $N_{cls}$ is set to be 32 to avoid sample bias as mentioned above, and $k$ is set to be 8 to maintain a mini-batch number of 256 since it does not affect our results but only the computation cost.

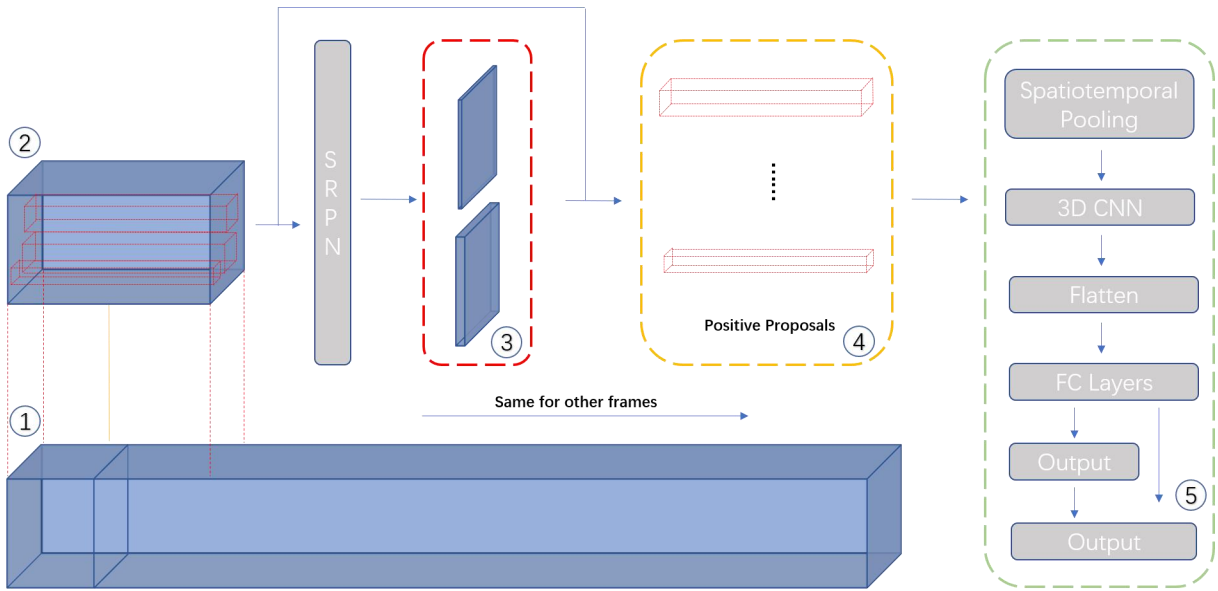

FIGURE 4: Procedure of classification. (1) Feature map of a batch, having a size of $L \times H/8 \times W/8 \times C$. (2) Bout features for classifying behaviors in frame $k + 1$, containing features of frame 1 to $2k + 1$. Likewise, the last bout in the batch is used for classifying behaviors in frame $L-k$, containing features frame $L-2k$ to frame $L$. Thus, there are $L-2k$ bouts in a batch of input. (3) Output of SRPN consist of a "score map" and a "coordinates map", stands for the probability of "behaviorness score" and adjusted bounding boxes coordinates of the corresponding anchor box. (4) Proposed bounding boxes are then sorted according to their "behaviorness score", $N_{cls}$ bounding boxes with highest score are selected and then fed into classifier. (5) The classifier starts with spatiotemporal pooling layer to pool features in bounding boxes into a fixed size followed by 3D convolutional layers and full-connected layers. The system first output a predicted parent category and then the use the output as well as output of full-connected layers for behavior classification.

The detection layers utilize the output of SRPN to locate features that may contains a behavior. At each bout, anchors boxes are sorted according to their probability of containing a behavior. Spatiotemporal features extracted by base layers are then selected by anchor boxes with highest probability of containing a behavior and fed into spatial region of interest layer and pooled into a fixed-size feature map for classification.

Denote the size of a proposed bounding box as $L \times W \times H$, the output size of SRoI pooling layer is $L \times W^0 \times L^0$. Where output of full-connected layers for behavior classification. $W$, $W^0$ stands for the input width and output width of Spatial RoI pooling layer(likewise for $L$, $L^0$ and $H$, $H^0$).

To avoid information loss, we first train models with fixed temporal length so that only the spatial size of feature map is changed in the SRoI pooling layer. The SRoI pooling layer works like RoI pooling layer at frame level. It dividing the $W \times L$ features at each frame into an $W^0 \times L^0$ grids, each grid then has an approximate size of $W/W^0 \times L/L^0$, and then max-pooling the values in each sub-window into the corresponding grid cell. SRoI then concatenate features at each frame and channel, so that the temporal information are kept unchanged. In our experiments, we train models with different temporal length and compared their speed and accuracy. Notice that our system applied classification at frame-level, the temporal features are only used for obtain a continuousness of an action to improve the accuracy of classification. Models with different temporal length are used to compare the effect of temporal information instead of locating action at temporal dimension.

### 3) Social behavior detection

Since proposals with different spatial size are pooled into fixed-size by SRoI pooling layers, we can adopt 3D CNN layers and 3D max-pooling for capturing temporal features and full-connected layers for classification.

Unlike most object detection and action detection tasks, different categories of behavior have high similarities in our tasks. However, there are still significant difference between social behavior and individual behavior. Thus, we adopt a two-step classification. On the first step, the classifier is trained for detecting individual/social behavior and the second-step classifier is in charge of detecting the specific category of behaviors. The output of super-category classifier feed back to the second classifier to reduce the confusion rate of behaviors and individual behaviors. In our experiments, we compare model with and without super-category classier.

As shown in Fig 4, both classifiers sharing the same fully connected layers. A model with super-category classifier first output a prediction of whether the proposal containing a individual or social behavior, and feed the prediction back into the classifier. The loss function of behavior detector is defined as:

$$L_{det} = \frac{1}{N_b} \sum_m (\frac{1}{N_{pro}} \sum_j L_{beh}(pro_i^m, pro_i^{m*})) \quad (5)$$

where $L_{det}$ is the loss of detector; $N_b$, is the number of predicted frames; $N_{pro}$ is number of proposals that fed into a detector per frame. $L_{beh}$ is the loss function of behavior detector, which is a categorical cross entropy loss function, $pro_i^m$ is the prediction of $i$th proposal in frame $m$ and $pro_i^{m*}$ is the ground true label of $i$th proposal in frame $m$.

## IV. EXPERIMENTS
### A. BASE LAYERS

We train and test our models on of fly-vs-fly dataset [7]. The fly-vs-fly dataset contains a total of 22 hours of 47 pairs of fruit flies interacting. The dataset contains three subsets: Boy meets boy is designed for study the sequence of actions between two male flies; Aggression is used to quantify the effect of genetic manipulation on their behavior; Courtship contains a female and a male and was used to study how genetic manipulation affects male courtship behavior.

We use three behaviors, namely: wing threat, hold, tussle, in Aggression dataset for training and detecting since behaviors in Aggression dataset are in a wider range and each class takes up approximately equal frames.

For training and testing, we extract labelled bout as video clip. To improve the effect of our detector, we also extend random frames at each side of the bout so that each video clip may contain normal behaviors before and after labelled actions. We also randomly pick 150 video clips in which none of listed behaviors are included to maintain a sample balance between normal action and labelled social behavior. Thus, there are 150, 132, 205, 146 videos, with length in range of 32 to 128 frames, for normal, containing wing threat, hold and tussle behaviors. The training, validation, and testing are 70%, 20% and 10% of the video clips that are randomly chosen from the whole dataset. Notice that since our system use a video clip as an input but outputs predictions at frame level, the exact steps of training, validation, testing steps are not fairly 7:2:1, but some proportion close to that.

### B. TRAINING

We train models with different base layers (Table 2) and different detectors (Table 3). All models are trained on a Nvidia GTX 1060 6GB GPU. Joint training is not available for such a complex network due to the constraints of memory; thus, we adopt alternating training algorithm.

TABLE 3: Experimental results with different base layers

| Models | Bout Length | Overall mAP | Fly | Hold | Tussle | Wing | Training Time(s/frame) | Testing Time(s/frame) |
|---|---|---|---|---|---|---|---|---|
| VGG | 1 | 33.6 | 28.1 | 36.7 | 44.9 | 14.6 | 0.29 | 0.042 |
| | 3 | 59.0 | 74.4 | 49.7 | 57.2 | 55.0 | 0.55 | 0.102 |
| | 5 | **63.7** | **75.2** | 46.0 | 65.5 | **67.9** | 0.86 | 0.168 |
| | 7 | 59.0 | 52.0 | **55.8** | **65.9** | 52.2 | 1.16 | 0.231 |
| ResNet | 1 | 37.1 | 34.2 | 40.7 | 47.1 | 26.4 | 0.20 | 0.037 |
| | 3 | 56.3 | 73.1 | 37.5 | 59.4 | 52.8 | 0.37 | 0.082 |
| | 5 | **62.1** | **75.0** | 44.9 | **68.3** | **70.2** | 0.53 | 0.121 |
| | 7 | 61.7 | 52.3 | **59.3** | 68.1 | 61.1 | 0.68 | 0.147 |

TABLE 4: Experimental results with different detectors

| Models | Bout Length | Overall mAP | Normal | Hold | Tussle | Wing | Training Time(s/frame) | Testing Time(s/frame) |
|---|---|---|---|---|---|---|---|---|
| VGG + 3C3D | 5 | **63.7** | 75.2 | 46.0 | **65.5** | **67.9** | 0.86 | 0.168 |
| VGG + 6C3D | 5 | 58.4 | 71.9 | 49.4 | 60.3 | 63.7 | 0.97 | 0.182 |
| VGG + 3C3D + S | 5 | 60.3 | **81.4** | 50.4 | 43.2 | 66.2 | 1.12 | 0.172 |
| VGG + 6C3D + S | 5 | 61.4 | 75.2 | **52.6** | 61.5 | 60.3 | 1.29 | 0.195 |

TABLE 5: Experimental results of different models

| Models | mAP | Fly | Hold | Tussle | Wing |
|---|---|---|---|---|---|
| Ours with VGG5 | **63.7** | **75.2** | 46.0 | 65.5 | 67.9 |
| Ours with ResNet | 62.1 | 75.0 | 44.9 | **68.3** | **70.2** |
| Faster RCNN [18] | 29.9 | 26.4 | 30.2 | 47.7 | 16.8 |
| Yolo [20] | 22.4 | 30.1 | 27.9 | 42.9 | 18.3 |
| C3D [24] | 56.8 | 68.7 | **49.2** | 56.9 | 51.5 |
| C3D + LSTM [25] | 48.5 | 55.7 | 37.9 | 50.0 | - |

TABLE 6: Experimental results of *ANK* vs IoU

| Model | Measure | Overall mAP | Fly | Hold | Tussle | Wing |
|---|---|---|---|---|---|---|
| VGG5 | ANK | **63.7** | 75.2 | **46.0** | **65.5** | 67.9 |
| | IoU | 62.0 | **77.3** | 41.4 | 59.7 | **70.3** |
| Res5 | ANK | **62.1** | **75.0** | **44.9** | **68.3** | **70.2** |
| | IoU | 59.7 | 71.4 | 38.9 | 63.3 | 67.5 |

At the very beginning of training, a SRPN with base layers is initialized with pre-trained model. Since we adopt modified pre-trained models, there are some layers in base layers different from that of the original model. Those layers, as well as the SRPN, are initialized by drawing weights from a zero-mean Gaussian distribution with standard deviation of 0.01. For each input video clip with length of $L$, the model is fine-tuned end-to-end. Weights and proposals with high "action scores" generated by SRPN are saved for detector training. Then a detector with base layers are then initialized by the saved weights. Unique layers in the detectors are initialized by drawing weights like that of SPRN. The input frames and the proposals generated in SPRN are fed into the detector for training. For the following batches, the training is processed in a similar way, but layers are initialized from the saved weights instead of pre-trained model. Thus, the SRPN and detector are sharing the same base layers.

We set the length of input($L$) to be $10 + 2k$ to keep a fixed training batch size. To avoid over-fitting, video clips of different behaviors are fed into the system alternatively. Each

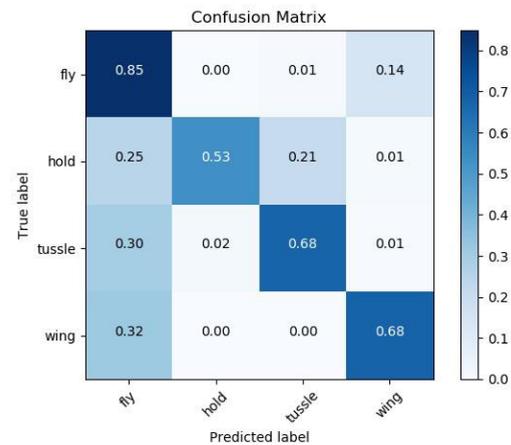

*A. (a) VGG + C3D*

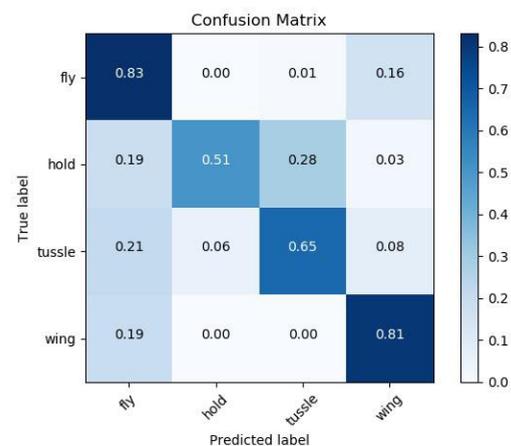

*B. (b) VGG + C3D + S*

FIGURE 5: Confusion matrix of models with and without super-category classifier.

model iterates for 25 epochs, each contains 80 batches, that is 20k frame-wise training in total. Since different behaviors are confusable, we set a relatively low learning rate for both SRPN and detector, $10^{-6}$, to avoid falling into local minima.

### B. EXPERIMENT RESULTS

We train models with 4 different bout length($k$) and 2 different 3D CNN architectures (Table 2) to compare the effects caused by bout length and 3D CNN structures.

In Table 3 reports the results of models with different $k$. The implementation details are presented in previous section and all models are trained and tested using same data segmentation.

**Model Performance**. Models with VGG16 base layers and a bout length of 5 achieves the top result on the dataset with a mAP of 63.7%. The model with $K = 1$, is a degenerated model that similar to an object detection model, has the lowest overall mAP as well as category mAPs. When we add spatiotemporal feature into our model, the result is significantly improved. "Normal" and "Wing threat" behaviors achieves a mAP of 75.2% and 67.9% in model with $k = 5$. However, when the bout length comes to 7, there is a slight decline in mAP which is caused by the accuracy decrease of Normal and Wing threat behavior. The result indicates that longer bout lengths cannot always improve the result of the detector, since more irrelevant frames are taking into consideration may result in a confusion. In contrast, longer bout length can improve the accuracy of hold and tussle behaviors, which are the most confusing behaviors in our task. The "Hold" and "Tussle" behaviors achieves its highest mAP in the model with bout length of 7, which are 55.8% and 65.9%, respectively.

At the same time, the running time of our models also increases with the increase of bout length since the computation cost of detector increase, even if we adopt a sliding window to reduce the cost of base layers. For training, the model with $K = 1$ has a lowest running time, which is $2.9s$ per batch, which is $0.29s$ per frame, and the model with $k = 7$ has a highest running time of $11.6s$ per batch, which is $0.116s$ per frame. For predicting, running time of model with $K = 1$ and $K = 7$ are $0.045s$ per frame and $0.231s$ per frame.

The ResNet based model, compared to that of VGG model, shows more stability when the bout length increases. With the bout length increase, the mAP of different models does not increase or decrease rapidly like that of VGG. On the other hand, the overall results are similar with that of VGG. The model with bout length of 5 has the highest overall mAP of 62.1%, as well as "Normal", "Tussle" and "Wing threat" behaviors, which are 75.0%, 68.3% and 70.2% respectively. Meanwhile, since ResNet has fewer parameters than VGG, the running time of training and testing is lower than that of VGG.

The result shows that detecting social behaviors with spatiotemporal features is effective. However, it is not always useful since reluctant features might result in confusion for short duration behaviors.

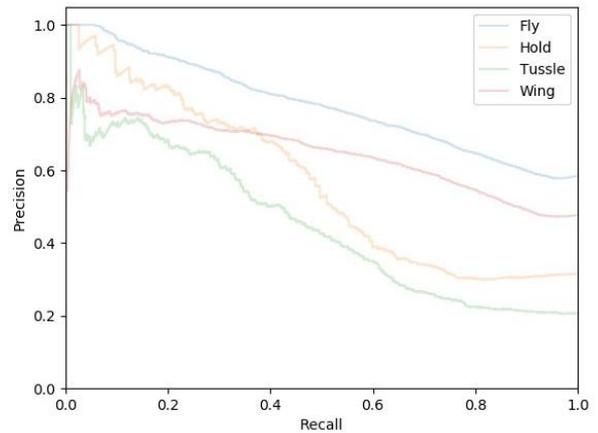

C. (a) VGG + C3D

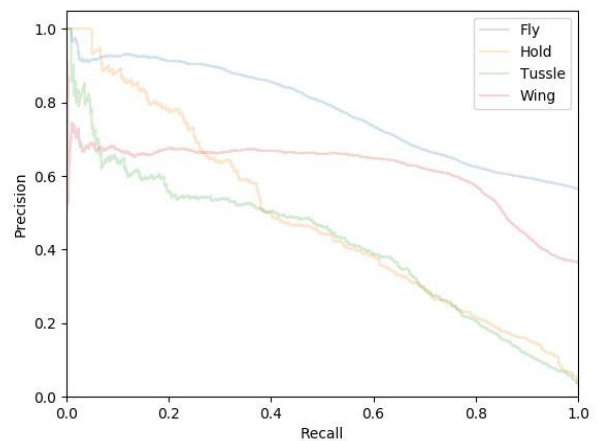

D. (b) VGG + C3D + S

FIGURE 6: Precision-Recall curve of model with and without super-category classifier.

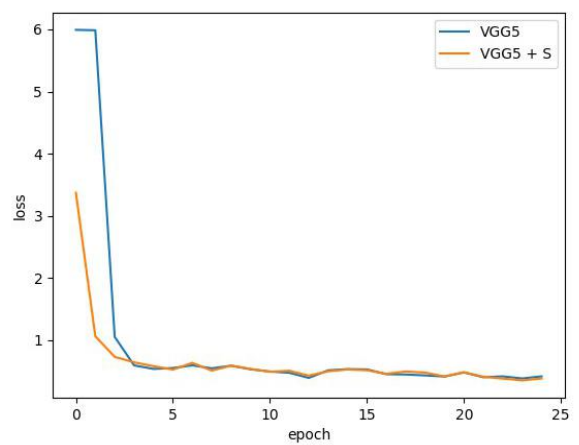

FIGURE 7: The blue and orange lines are training loss of VGG5 and VGG5 + S, respectively.

**Model architectures.** In Table 4, we compare the mAP of models with different detector architectures. Fixing bout length and base layers of our model, we compare the effect caused by 3D CNN network and super-category detector. VGG + 3C3D has the highest Overall mAP of 63.7%. This is higher than the VGG +6C3D (58.4%) by 5.3%. The result shows that a deeper structure does not always have better performance.

Another comparison in Table 4 is between models with and without super-category classifier. It was a surprise to us as the model with super-category classifiers does not outperform the one without. The VGG + 3C3D + S has a mAP of 60.3%, which is lower than VGG + 3C3D by 3.4%. Figure 5 shows the confusion matrix of VGG + 3C3D and VGG + 3C3D + S. The false positive rate of a social behavior (hold, tussle), to be recognized as single behavior (fly, wing threat) reduces slightly (i.e. "hold" towards fly is 0.25 in Fig.5-a) and reduced by 0.06 in Fig.5-b)). However, it is more likely to be recognized as another social behavior (i.e. hold towards tussle in (a) is 0.21 but increases to 0.28 in (b)). Fig.6 shows the Precision-Recall curve of VGG + C3D and VGG + C3D + S. The curve of tussle and hold behaviors in VGG + C3D are more stable than that of in VGG + C3D + S.

Fig.7 compares the training loss of VGG +3C3D and VGG + 3C3D + S. Both losses are composed of *cls* loss, reg loss of SRPN and *cls* loss of classifier. It indicates that the super-category classifier can boost the training process but also makes it unstable.

**Comparison between models.** Table 5 shows the mAP of SRCNN compared with popular object detection and action detection modules. Since the detection is performed frame-wise, we compared the result of our model with object detection method as well as action detection methods. The faster RCNN and Yolo are trained and tested with single frame, while the C3D and C3D + LSTM output a prediction of current frame using few frames that centered on the current frame. In the "Wing Threat" dataset, there might be more than one behavior in a single frame. To avoid confusion, we remove the dataset when training and testing C3D + LSTM model.

The result shows that our system over-perform both object detection task since we take temporal features into account. On the other hand, our model is better than action detection methods since our feature extraction module is carefully designed to avoid any spatial feature loss. The highest overall mAP comes from VGG5 based model of our SRCNN, followed by a C3D based model with mAP of 61.6%, which shows that feature extraction with 2D CNN instead of 3D CNN can slightly increase the performance of our model.

*ANK* **vs IoU.** Table 6 compares the performance between models trained with anchors that are labelled by *ANK* and IoU measure. It is intuitive that adopting IoU as the bounding tube measure might result in a decline in classification ability of our model since a bounding box might have high IoU with object even only part of the object is contained in the bounding box, which might be a problem for confusable behaviors detection (Fig.1). To address the issue, we introduced *ANK* measure for setting labels of anchor tubes. The mAP of VGG5 + *ANK* is 1.7% higher than VGG5 + IoU. For the Res based model, the gap is 2.4%. The result shows that the *ANK* measurement we proposed for setting ground truth box is more effective for confusable behavior detection.

## V. CONCLUSIONS AND FUTURE WORK

The purpose of this paper is to develop a deep neural network system for complex social behavior detection low resolution videos. The classification is performed frame-wise. To take temporal information into account, the system starts with a 2D CNN for feature extraction, followed by 3D CNN for spatiotemporal feature fusion, spatial region proposal generation and classification. We adopt a sliding window at the data pre-processing step and use time distributed 2D CNN layers to reduce computation cost. Meanwhile, we modify the architecture of the pre-trained model to maintain a balance between additional training cost and spatial information loss. In the classification layers, we first propose a SPRN (spatiotemporal region proposal network) to generate feature tube. We also proposed a new measurement for setting ground truth label of tube proposals since IoU might be ineffective for confusable behaviors among laboratory animals. The spatiotemporal pooling layer pools tube proposals into tubes with fixed length, width and height. Finally, we introduce a super-category classifier to boost the training process.

The results of our work show that: 1) The 2D + 3D architecture has better performance than the 2D object detection methods as well as action detection methods that have been designed for human action detection; 2) Deeper 3D CNN architectures cannot always improve the performance of the model; 3) The *ANK* measure is more effective in labelling positive anchors than IoU in our task; 4) The super-category classifier can boost the training, but it makes the process unstable. These findings indicate that for confusable social behavior detection in low resolution video, the key step is not feature fusion but feature extraction, which means that the more original information from video is captured by the system, the better its performance will be.